\documentclass[10pt,a4paper,twocolumn]{article}
\usepackage[T1]{fontenc}
\usepackage{titlesec}
\usepackage{authblk}
\usepackage{color}
\usepackage{geometry}
\usepackage{setspace}
\usepackage[pdftex]{graphicx}
\usepackage{float}
\usepackage{caption}
\usepackage{amsmath}
\usepackage{cite}
\usepackage{enumitem}
\usepackage{doi}
\usepackage[colorlinks=false,linkcolor=black,citecolor=black,urlcolor=black]{}
\usepackage{hyperref}
\hypersetup{hidelinks}

\titlespacing*{\section}{0pt}{30pt}{0pt} 
\titlespacing*{\subsection}{0pt}{20pt}{0pt}
\titleformat{\section}{\Large\bfseries}{\arabic{section}.}{0.5em}{} 
\titleformat{\subsection}{\large\bfseries}{\thesection.\arabic{subsection}.}{0.5em}{}

\numberwithin{equation}{section}

\let\-=\boldsymbol

\begin{document}
\title{\textbf {ODEs learn to walk: ODE-Net based data-driven modeling for crowd dynamics}}
\author{Chen Cheng$^1$ and Jinglai Li$^2$ \thanks{j.li.10@bham.ac.uk}}
\affil{$^1$School of Mathematical Sciences, \protect\\ Shanghai Jiao Tong University, Shanghai 200240, China}
\affil{$^2$School of Mathematics, \protect\\ University of Birmingham, Edgbaston, Birmingham B15 2TT, UK}

\renewcommand*{\Affilfont}{\small\it} 
\renewcommand\Authands{ and } 
\date{}

\maketitle
\small

\begin{abstract}
Predicting the behaviors of pedestrian crowds is of critical importance for a variety of real-world problems. 
Data driven modeling, which aims to learn the mathematical models from observed data, is a promising tool to construct models 
that can make accurate predictions of such systems. 
In this work, we present a data-driven modeling approach based on
the ODE-Net framework, for constructing continuous-time  models of crowd dynamics. 
We discuss some challenging issues in applying the ODE-Net method to such problems, which are primarily associated with 
the dimensionality of the underlying  crowd system,
and we propose to address these issues by
incorporating the social-force concept in the ODE-Net framework. 
Finally application examples are provided to demonstrate the performance of the proposed method. 
\end{abstract}


\section{Introduction} 

Collective motion of pedestrians is a highly common phenomenon in  urban life, 
and understanding the dynamics of pedestrian crowds is essential for a large variety of applications, 
ranging from safety management~\cite{johansson2008crowd,Helbing2011} to robot navigation~\cite{trautman2010unfreezing,trautman2015robot}. 
Modeling the behaviors of pedestrian crowds has 
attracted considerable attention in multiple disciplines such as physics, social science and artificial intelligence,
and various models have been proposed in the past decades. 
Due to the complexity of the crowd dynamics, 
driving the mathematical models  
that can accurately predict the crowd behaviors is an extremely challenging task. 
To this end a particularly promising remedy is to develop mathematical models with 
the assistance of related data, 
an approach often referred to as data-driven modeling~\cite{Solomatine2008}. 

Within the context of crowd dynamics modeling, we here discuss two main strategies behind the data driven methods. 
The first strategy assumes that the crowd dynamics follows a specific mathematical model that is usually derived based on physics, 
but all or some of the model parameters are not available; one then estimates these parameters by
fitting the observation data into the model. 
Examples of such methods include \cite{johansson2007specification,moussaid2009experimental,tang2011approach,wolinski2014parameter},
among some others.
While this type of methods are conceptually  straightforward and relatively easy to implement, their performance is ultimately limited
by the mathematical models adopted. 
The second strategy offers more flexibility: namely 
it does not impose a specific mathematical model;
rather, it learns the model (often represented by an artificial neural network) directly from the data with machine learning techniques. 
While their implementation is usually more complicated,
the machine-learning based methods are much less restrictive than the first kind
and can potentially obtain very accurate model, provided that high-quality  data are available.

In the past a few years, various efforts have been made to the machine learning based
data driven modeling, e.g., \cite{yi2016pedestrian,alahi2016social,amirian2019data,antonucci2020generating}. 
To the best of our knowledge, most of these existing  methods are designed to learn
crowd dynamics models that are discrete in time, 
largely because the discrete-time models can be naturally formulated with a deep neural network such
as the recurrent neural network (RNN).
On the other hand, there is strong desire to develop continuous-time models,
as they can be used to predict the crowd behaviors at any time of interest. 
The ODE-Net method, first proposed in \cite{chen2018neural}, has gained attention as a tool to learn continuous-time models of physical systems~\cite{chen2020learning,zhong2019symplectic,rubanova1907latent,huang2020learning}.
Simply speaking, ODE-Net formulates the system of interest as an ordinary differential equation system, 
which is represented 
by a deep neural network, and learned from the data. 
The ODE-Net  method, however, can not be directly applied to the crowd dynamics, 
and we summarise three main challenges of it, all associated with the crowd size (or equivalently the dimensionality of the system): 
first and foremost, due to the high training cost, ODE-Net 
generally has difficulty dealing with systems of high dimensions,
rendering it especially unsuited for large-size crowds;
secondly, in reality the size of a crowd may vary in time, with pedestrians entering or leaving the scene of interest,
and such a system can not be easily modeled by ODE-Net; 
finally, the model obtained by ODE-Net cannot be used to predict crowds whose size is different from the training system, which makes its application very limited.  
In this work we propose an ODE-Net based method to learn the crowd dynamics models from data,
where the aforementioned issues are addressed by 
incorporating underlying physical knowledge of the dynamics into the  ODE-Net model.
In particular, we adopt the concept that the crowd is a physical system 
driven by social and psychological forces as is in the so-called 
social force model (SFM)~\cite{Helbing2000VICSEK}, and then learn
those force functions from data.
The resulting social force based method allows one to learn the models from data for 
large-scale and variable-size crowds,
and also use the learned models to predict the behaviors of crowds of any sizes. 

The rest of the paper is organized as follows. 
In Section~\ref{sec:method} we present the social force based ODE-Net method,
and in Section~\ref{sec:examples} we demonstrate the performance of the proposed method by
applying it to  data generated from 
two commonly used crowd dynamics models. 
Finally Section~\ref{sec:conclusion} offers some conclusions and discussions.

\section{Methodologies} \label{sec:method}

\subsection{ODE-Net for crowd dynamics}\label{sec:odenet}
We start by introducing the ODE-Net from a deep neural network perspective. 
Traditional deep neural networks, such as residual networks, build complicated transformations by composing a sequence of transformations to a hidden state:
\begin{equation}\label{eq:resnet}
\frac{\-z_{t+\delta_{t}}-\-z_{t}}{\delta_{t}}=h_{t}\left(\-z_{t}\right), \quad \delta_{t}=1,
\end{equation}
where $h_{t}\left(\-z_{t}\right)$ is a function parameterized by a neural network. These iterative updates can be interpreted as an Euler discretization of a continuous transformation. In contrast to traditional deep neural networks where $\delta_{t}=1$ is fixed, ODE-Net~\cite{chen2018neural} introduced a continuous version in which $\delta_{t} \rightarrow 0$. 
As a result, Eq.~\eqref{eq:resnet} becomes
\begin{equation}\label{eq:odenet}
\frac{\mathrm{d} \-z(t)}{\mathrm{d} t}=h(\-z(t), t).
\end{equation}
In this continuous framework, training the networks becomes to learn the function $h(\-z,t)$
and next we will discuss how to learn this function. 

First we assume that the function $h(\-z,t)$ is represented by a neural network $h_{\theta}(\-z,t)$ parameterized by $\theta$,
and we have observed data at $t_0$ and $t_1$, denoted as $\hat{\-z}(t_0)$ and $\hat{\-z}(t_1)$ respectively. 
Starting from the input layer $\hat{\-z}(t_0)$, the output layer $\-z(t_1)$ can be defined by the solution to this ODE initial value problem at some time $t_1$:
\begin{equation}
\-z(t_1)=\hat{\-z}(t_0) + \int_{t_0}^{t_1}h_\theta(\-z(t), t) dt, \label{e:odeint}
\end{equation}
and the time from $t_0$ to $t_1$ is referred to as the integration time of the data point. 
Eq.~\eqref{e:odeint} can be computed using an off-the-shelf differential equation solver and we write it as,
\begin{equation}
{\-z}(t_1)=\operatorname{ODESolve}\left(\hat{\-z}(t_0), h_{\theta}, t_0, t_1\right).
\end{equation}
The network parameters $\theta$ are computed by iteratively minimizing a prescribed loss function $L(\hat{\-z}(t_1), \-z(t_1))$, which measures the difference between the observed data $\hat{\-z}(t_1)$ and the model prediction $\-z(t_1)$. 
An interesting feature of this method is that the gradient of the loss function with respect to $\theta$ 
can be computed using the adjoint sensitivity method, which is more memory efficient than directly backpropagating through the integrator~\cite{chen2018neural}.

As has been discussed earlier, 
ODE-Net allows us to construct a continuous-time model for the crowd dynamics. 
Namely, let $\-z(t)$ represents the state of the crowd at time $t$ and as a result Eq.~\eqref{eq:odenet} becomes the governing equation of the crowd dynamics; suppose that we have observations of the crowd flow $\hat{\-z}(t)$, and we can use the training process described above to 
learn the function $h(\-z,t)$ (or more precisely its neural network representation $h_\theta(\-z,t)$). 

Though the application of ODE-Net to crowd dynamics is conceptually straightforward, 
the implementation is highly challenging. 
When applied to crowd dynamics, $\-z$ represents the state of motion of the entire crowd that may consist of a large number of particles (i.e., pedestrians, and throughout the paper we use these two terms interchangeably),
and it follows that $\-z$ can be of very high dimensions since the dimensionality of $\-z$ is proportional to the size of the crowd. 
In this case, learning a high-dimensional function $h(\-z,t)$ can be prohibitively difficult:  
it may require a massive amount of training data which may not be available in practice, and 
the computational cost for training such a complex model can be exceedingly high. 
In addition, as one can see, in the formulation described above,  the dimensionality of $\-z$ needs to be fixed, 
which often does not meet the reality, as in most situations people may enter or leave the scene of interest and the dimensionality of $\-z$ varies over time. 
More importantly, as the dimensionality of $\-z$ is fixed, once the model is learned from the data, 
it can only be used to predict systems of the same number of particles, a serious limitation of the usefulness of the method. 
To address these issues, we propose to address the dimensionality issue by incorporating the social force (SF) concept into 
the ODE-Net method, which is detailed in Section~\ref{sec:sfm}. 

\subsection{Social-force based ODE-Net} \label{sec:sfm}

Suppose that we consider a crowd of $N$ particles and we can write the state variable $\-z = (z_1,...,z_N)^T$ 
where $z_n$ represents the state of motion of particle $n$ for each $n=1...N$.
In particular we have $z_n= (x_n,v_n)$  where $x_n$ and $v_n$ are respectively the position and the velocity of particle $n$. 
We also introduce the notations $\-x=(x_1,...,x_N)^T$ and $\-v=(v_1,...,v_N)^T$. 
Now according to the Newton's second law, model~\eqref{eq:odenet} can be re-written as
\begin{equation}\label{e:newton}
\left[\begin{array}{l}
\dot{\boldsymbol{x}} \\
\dot{\boldsymbol{v}}
\end{array}\right]=\left[\begin{array}{c}
\boldsymbol{v} \\
\mathcal{M}^{-1}\-f
\end{array}\right],
\end{equation}
where $$
\-f(\-x,\-v) = 
\begin{pmatrix}
f_1(\-x,\-v)\\...\\f_N(\-x_,\-v)
\end{pmatrix}$$ with $f_n(\-x,\-v)$ being the force applied to particle $n$ and 
$\mathcal{M} = \mathrm{diag}[m_1,...,m_N]$ with $m_n$ being the ``mass'' of particle $n$. 
With formulation~\eqref{e:newton}, the original ODE-Net problem is turned into  learning 
the force function $\-f(\-x,\-v)$ and estimating the mass matrix $\mathcal{M}$,
where one can see that learning function  $\-f(\-x,\-v)$ is by far the more challenging task here. 

\begin{figure*}[!htb]
\centering
\includegraphics[width=0.9\textwidth]{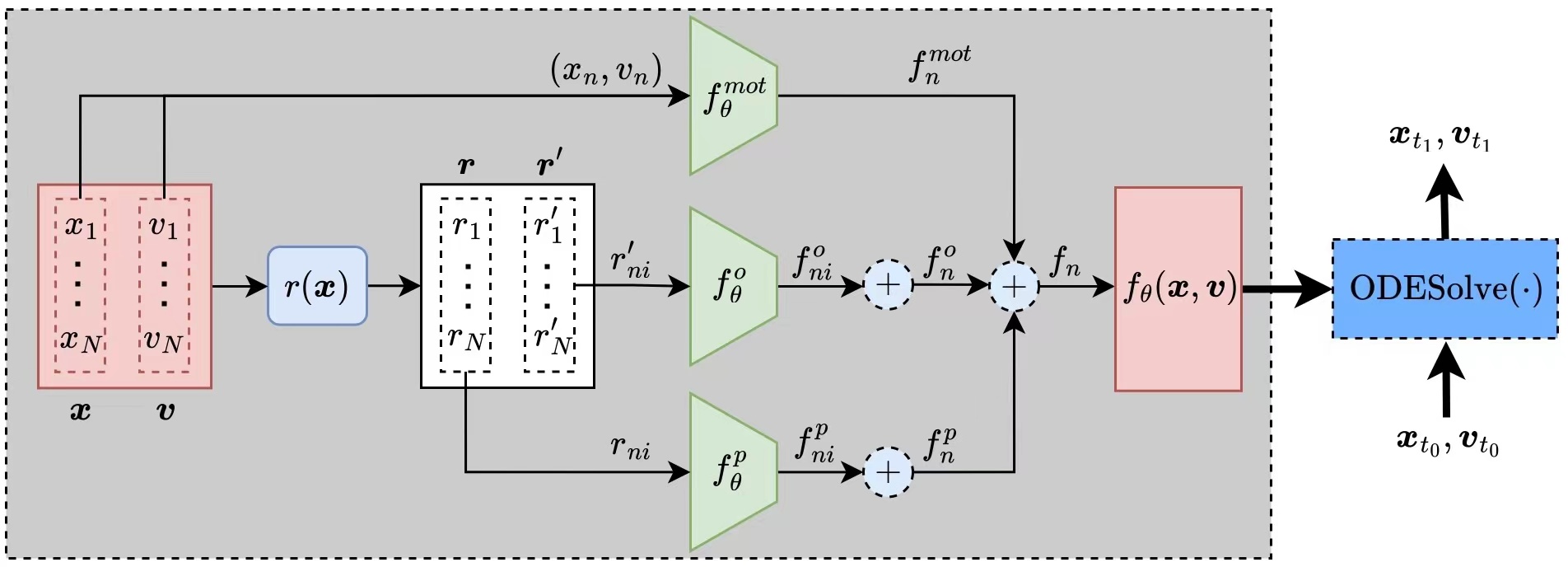}
\caption{Graphic representation of the SF-based ODE-Net.
The shaded box shows that how the social force is computed. 
The ODE system defined by the force via Eq.~\eqref{eq:augment}, 
enters the ODE solver together with the initial states~$\boldsymbol{x}_{t_0},\,\boldsymbol{v}_{t_0}$, yielding predictions of the states $\boldsymbol{x}_{t_1},\,\boldsymbol{v}_{t_1}$.}
\label{fig.graph}
\end{figure*}

It is important to note that in such problems $\-f$ and $\mathcal{M}$ should not be understood as the usual physical forces and masses respectively.
Rather, following the assumption of the  social force model~\cite{Helbing2000VICSEK}, 
$\-f$ represents the socio-psychological forces 
 driven by personal motivations and environmental constraints,
 and the mass matrix $\mathcal{M}$ characterizes how easy or difficult 
 to change the velocity of each pedestrian. 
  At this point the force field $\-f(\-x,\-v)$ is still a high dimensional function for large crowd size $N$,
 and further simplification is needed to make the learning problem feasible.

 We now introduce further assumptions to simplify the force function. 
 First we assume that the total force applied to each particle/pedestrian consists of two parts:
 \begin{equation}
 f_n = f^{mot}_n + f^{int}_n,\label{e:fn}
 \end{equation}
 where $f_n^{mot}$ is the force generated by personal motivation to 
 reach certain desired state of motion,
 and $f_n^{int}$ is the force caused by the interactions 
 with other particles and the environments (e.g., obstacles).  
 The total interaction force is further written as,
\begin{equation} \label{eq:sfm2}
{f}_n^{int}=\sum_{j(\neq n)=1}^N {f}^p_{nj}+\sum_{w=1}^W {f}^o_{nw},
\end{equation}
 where ${f}^p_{nj}$ is the interaction force between pedestrians $n$ and $j$ and ${f}^o_{nw}$ between 
pedestrian $n$ and the $w$-th obstacles (assuming there are $W$ obstacles in total). We now need to deal with both the motivation and the interaction forces. 
We first assume that the personal motivation force depends on the particle's state of motion:
\begin{equation}
f^{mot}_n = f^{mot}_\theta(x_n,v_n,d), \label{e:fmot}
\end{equation}
 where $d$ represents some environmental factors that also affect the motivation force, and $f^{mot}_\theta(\cdot)$ is an artificial neural network parametrized by $\theta$.   
 Next we consider the interaction force $f^{int}$.
To this end, it is common to assume that pedestrians psychologically tend to keep a distance between each other and avoid hitting obstacles.
 As such, the two interacting forces can be written as, 
 \begin{subequations}\label{e:fpfo}
 \begin{align}
& f^p_{nj} = f^p_\theta(r_{nj},u_{nj}),\\
 & f^o_{nw} = f^o_\theta(r_{nw},u_{iw}),
 \end{align}
 \end{subequations}
 where $r_{nj} = x_j-x_n$ is the relative location of particle $j$ to particle $n$,
  and $u_{nj} = v_j-v_n$ is the relative velocity of particle $j$ to particle $n$, and $r_{nw}$ and $u_{nw}$
  are defined in the same way for the obstacles.  Note here that  $v_w$ is usually $0$ in practice.

Under these assumptions, the total force function is completely determined by $f_\theta^{mot}$, $f_\theta^p$ and $f_\theta^{o}$.
\emph{Importantly the dimensionality of these three functions is independent on the crowd size $N$,
and therefore the learning problem for constructing the ODE-Net model is of fixed dimensionality regardless of how large  the crowd is. }
Relatedly the resulting model can be applied to a crowd of any size once the functions are learned.

\subsection{Implementation of SF-based ODE-Net}\label{sec:imp_odenet}
In this section we discuss how to implement the SF based ODE-Net method for crowd dynamics.
Simply speaking, one just inserts the social force functions $f_\theta^{mot}$, $f^p_\theta$ and $f_\theta^{o}$
 into the ODE model~\eqref{e:newton} via Eqs.~\eqref{e:fn}--\eqref{e:fpfo},
 and then trains the resulting ODE-Net with the algorithms described in Section~\ref{sec:odenet}. 
In what follows we provide further implementation details.

 In our numerical implementation in this work we make the following assumptions: 
 \begin{enumerate}[label=\alph*)]
 \item the masses of all particles are the same; this is of course a simplification, but it is important
 for the application of the learned model, as we typically do not have the knowledge  of the ``mass'' of each pedestrian when applying
 the learned ODE-Net model; 
 
\item a particle is only interacted with its $K$ nearest neighbors, a measure imposed to reduce the computational cost;
\item the motivation force of a particle only depends on the velocity and position of it 
and no environmental factors are explicitly included; %(but their information can be implicitly learned from from the data)
\item the interaction force between two particles (or a particle and an obstacle) only 
depends on the relative position of the objectives. 
\end{enumerate}
It is important to note that any of these assumptions can be removed or modified without affecting the implementation procedure described here --
for example
the interaction force can also depend on the relative velocity between particles. 

  The governing model~\eqref{e:newton} can be represented by a collection of the models for each particle in the form of: 
 \begin{equation}\label{eq:augment}
\left[\begin{array}{l}
\begin{bmatrix}
\dot{{x}}_1 \\
\dot{{v}}_1
\end{bmatrix}\\
...\\
\begin{bmatrix}
\dot{{x}}_N \\
\dot{{v}}_N
\end{bmatrix}
\end{array}\right]=\left[\begin{array}{c}
\begin{bmatrix}
{v}_1 \\
{f}_\theta({x}_1, {v}_1, r_1, r'_1) /{m_1}
\end{bmatrix}\\
...\\
\begin{bmatrix}
{v}_N \\
{f}_\theta({x}_N, {v}_N, r_N, r'_N) /{m_N}
\end{bmatrix}
\end{array}\right], 
\end{equation}
 where
 \begin{gather*}
 f_\theta({x}_n, {v}_n, r_n, r'_n) =f^{mot}_\theta(x_n, v_n) + f^p_{\theta}(r_n) + f^o_{\theta}(r'_n),\\
 r_n =(r_{n1},...,r_{nK}),\quad  r'_n = (r_{n1},...,r_{nW}).
 \end{gather*}
Here $r_n$ collects the relative locations of the $K$ nearest particles  to the $n$-th particle
 (i.e., $r_{ni} = x_i-x_n$),
and $r'_n$ are those of the $W$ obstacles, which can be easily computed by a deterministic function $r(\-x)$.
Note that in the present formulation, we omit the relative speed $u_n$ due to the experimental phenomenon that physical forces caused by bumping and body compression are infrequent in particle interactions. In other words, the interaction force is dominated by psychological forces, which only depend on $r_n$ according to the SFM.  Following \cite{zhong2019symplectic} we provide the graphic representation of the SF-based ODE-Net in Figure~\ref{fig.graph}. 

When training the ODE-Net model, all the collected data are organised in a way such that each data point consists of 
the states of the crowd at the initial and the end times:
$\{[(\hat{\-x}^i_{0},\hat{\-v}^i_{0}),(\hat{\-x}_{1}^i,\hat{\-v}_{1}^i)]\}_{i=1}^M$.
Each data point is plugged into the procedure described in Section~\ref{sec:odenet}
to update the force functions $f_\theta^{mot}$, $f_\theta^p$ and $f_\theta^{o}$.
It is worth mentioning that it is not necessary for all the data points to be collected for the same crowd;
rather we just need that  the crowd system remains the same within the integration time  
for each data point: namely, no particle enters or leaves the scene of interest from the initial time to the end time of a data point. 
We reinstate that this is possible thanks to the fact that all the  data points are used to update the same force functions. 

\section{Numerical experiments}\label{sec:examples}
\begin{figure}[!htb]
\centering
\includegraphics[width=.45\textwidth]{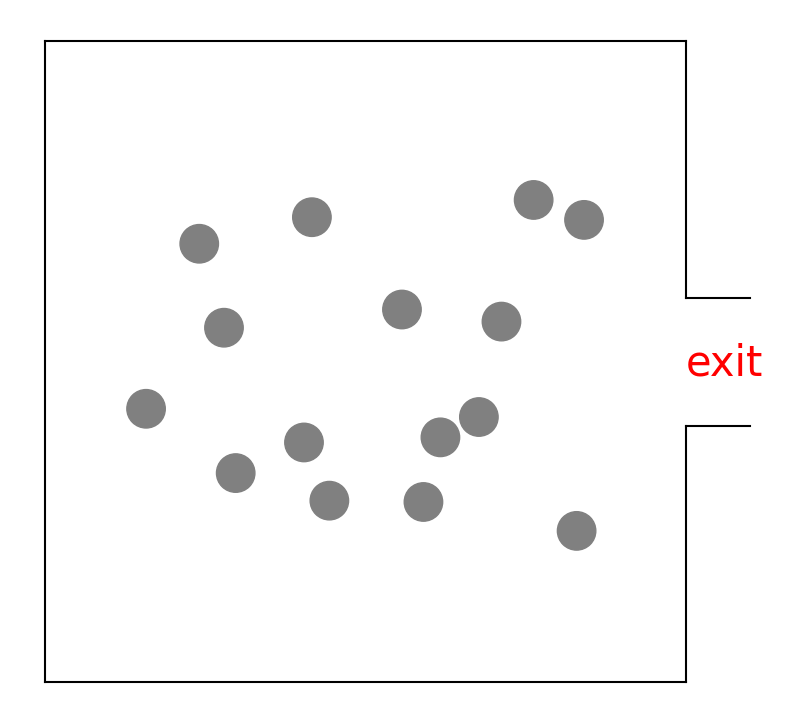}
\caption{Schematic plot of the simulation scenario. Pedestrians are depicted by dots and walls (i.e., obstacles) are depicted by lines. All pedestrians are moving towards the center of the exit.}
\label{fig.room}
\end{figure}

In this section, we 
conduct numerical experiments to demonstrate the performance of the proposed method. 
 Specifically we consider  a typical scenario where a crowd of individuals leave a room via a single exit, 
 as is shown in Fig.~\ref{fig.room},
 and synthetic data is used, which means that the true models are available for validating the results. 
 In our experiments we first generate data from a specific computer model under the aforementioned scenario, 
 and then learn the underlying model from the generated data using the proposed ODE-Net based method. 
 Finally the behavior of the learned model is compared with that of the true model to assess its performance. 
 
 We generate the  data with two representative models: one is continuous in time and the other is discrete.
 The continuous-time model adopted is the SFM mentioned earlier,
 and the discrete time one is the Optimal Reciprocal Collision Avoidance (ORCA) model~\cite{berg2011reciprocal},
 both of which are widely used in describing crowd dynamics. 

\subsection{Learning the social force model}
We first test with the  data generated from  SFM.  
The parameter values used in our simulation are presented in Table~\ref{tab.1}, largely following~\cite{Helbing2000VICSEK}. 
Training data is the simulated trajectories of 5 pedestrians, but as has been mentioned, the learned ODE-Net model can be applied to crowds of any size. 
The loss function used in all our experiments is
 $$L(\-x_{t_1}, \-v_{t_1}; \hat{\-x}_{t_1}, \hat{\-v}_{t_1}) = \|\-x_{t_1} - \hat{\-x}_{t_1}\|_1 + \|\-v_{t_1} - \hat{\-v}_{t_1}\|_1,$$ 
 where $\-x_{t_1}, \-v_{t_1}=\operatorname{ODESolve}(\-x_{t_0}, \-v_{t_0}, f_{\theta}(\-x, \-v), t_0, t_1)$
 are outputs of the ODE-Net. 
 The training process is terminated after 30 epochs.
Once the ODE model is learned, 
we  use it to predict the dynamics of a new crowd system, and compare the results with those of the actual model.

\begin{table}[!htb]
\small
\caption{List of parameter values in SFM.}
\label{tab.1}
\begin{tabular}{lll}
 \hline
 Variable&Value&Description \\
 \hline
  $H_0$ & $10\,\rm{m}$ & side length of the room\\
  $N$ & $5 (20)$ & number of pedestrians\\
  $m$ &  $80\,\rm{kg}$  & mass of pedestrians \\
  $v^p$ & $1.0\,\rm{m/s}$   & desired velocity \\
  $\tau$ & $0.5\,\rm{s}$  & acceleration time \\
  $r$ & $0.3\,\rm{m}$  & radius of pedestrians\\
  $A$ & $2\times 10^3\,\rm{N}$  & interaction strength\\
  $B$ & $0.08\,\rm{m}$ & interaction range\\
  $k$ & $1.2\times 10^5\,\rm{kg/s^2}$ & bump effect \\
  $\kappa$ & $2.4\times 10^5\,\rm{kg/(m\cdot s)}$ & friction effect \\
  $\Delta t$ & $0.001\,\rm{s}$ & time step in simulation\\
 \hline
\end{tabular}
\end{table}
 
\begin{figure*}[!htb]
\centering
\includegraphics[width=\textwidth]{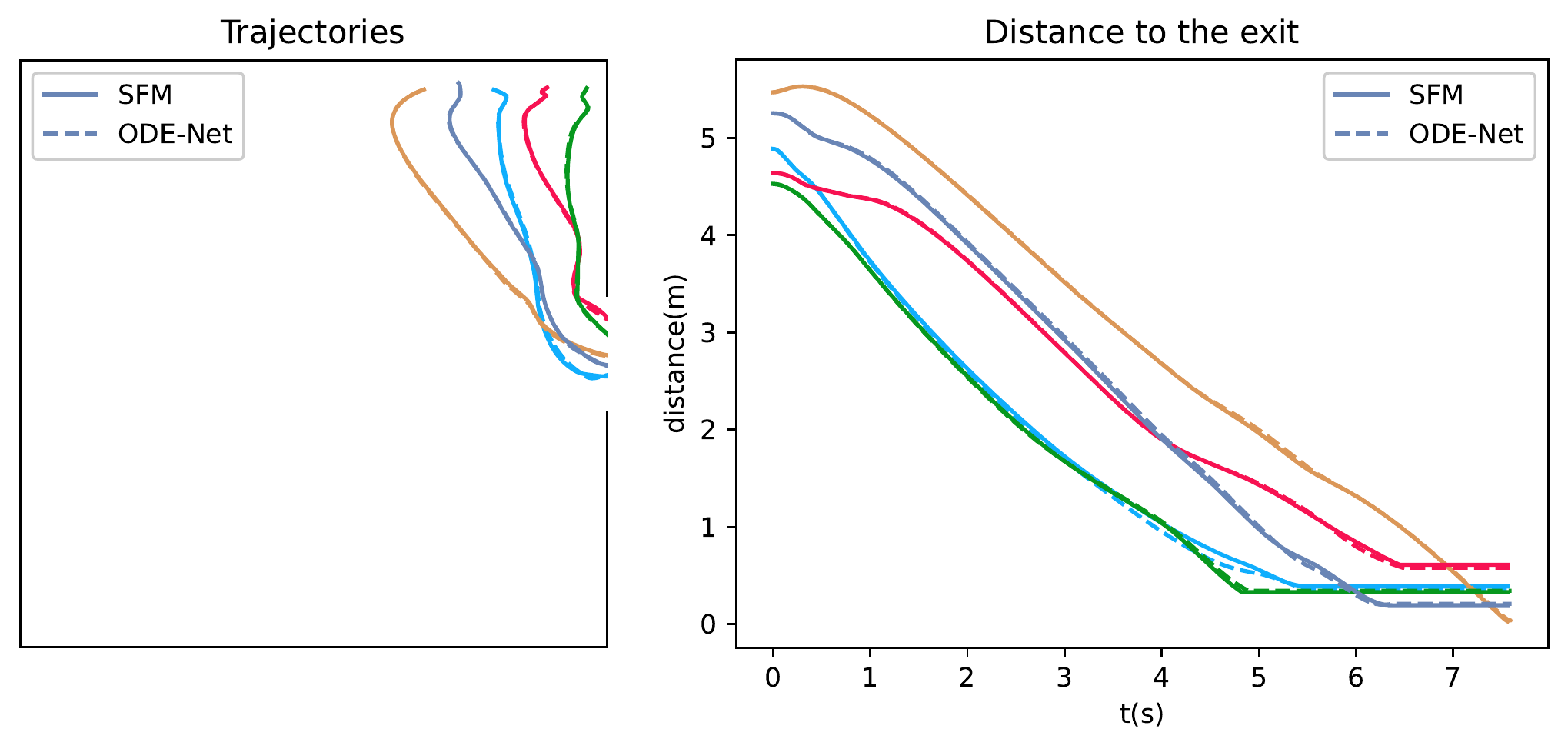}
\caption{Comparison of SFM and the learned ODE-Net model, for a crowd of 5 pedestrians. Left: pedestrian trajectories generated with SFM (solid) and ODE-Net (dotted). Right: each pedestrian's distance to the exit plotted against time,
where 
the results of SFM and of ODE-Net are shown as solid and dotted lines respectively.}
\label{fig.sfm5}
\end{figure*}
\begin{figure*}[!htb]
\centering
\includegraphics[width=\textwidth]{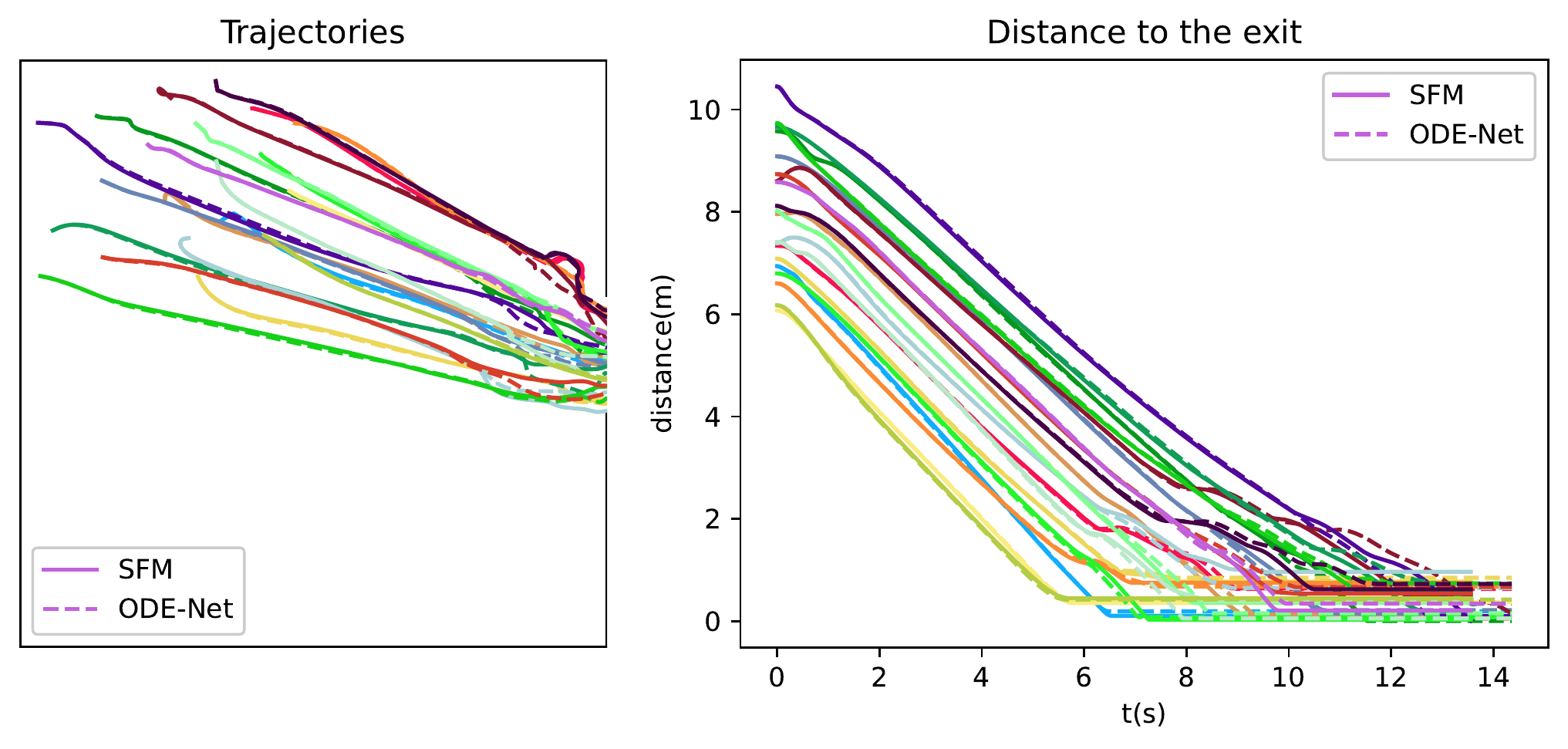}
\caption{Comparison of SFM and the learned ODE-Net model, for a crowd of 20 pedestrians. Left: pedestrian trajectories generated with SFM (solid) and ODE-Net (dotted). Right: each pedestrian's distance to the exit plotted against time, where 
the results of SFM and of ODE-Net are shown as solid and dotted lines respectively.}
\label{fig.sfm20}
\end{figure*}
\begin{figure*}[!htb]
\centering
\includegraphics[width=\textwidth]{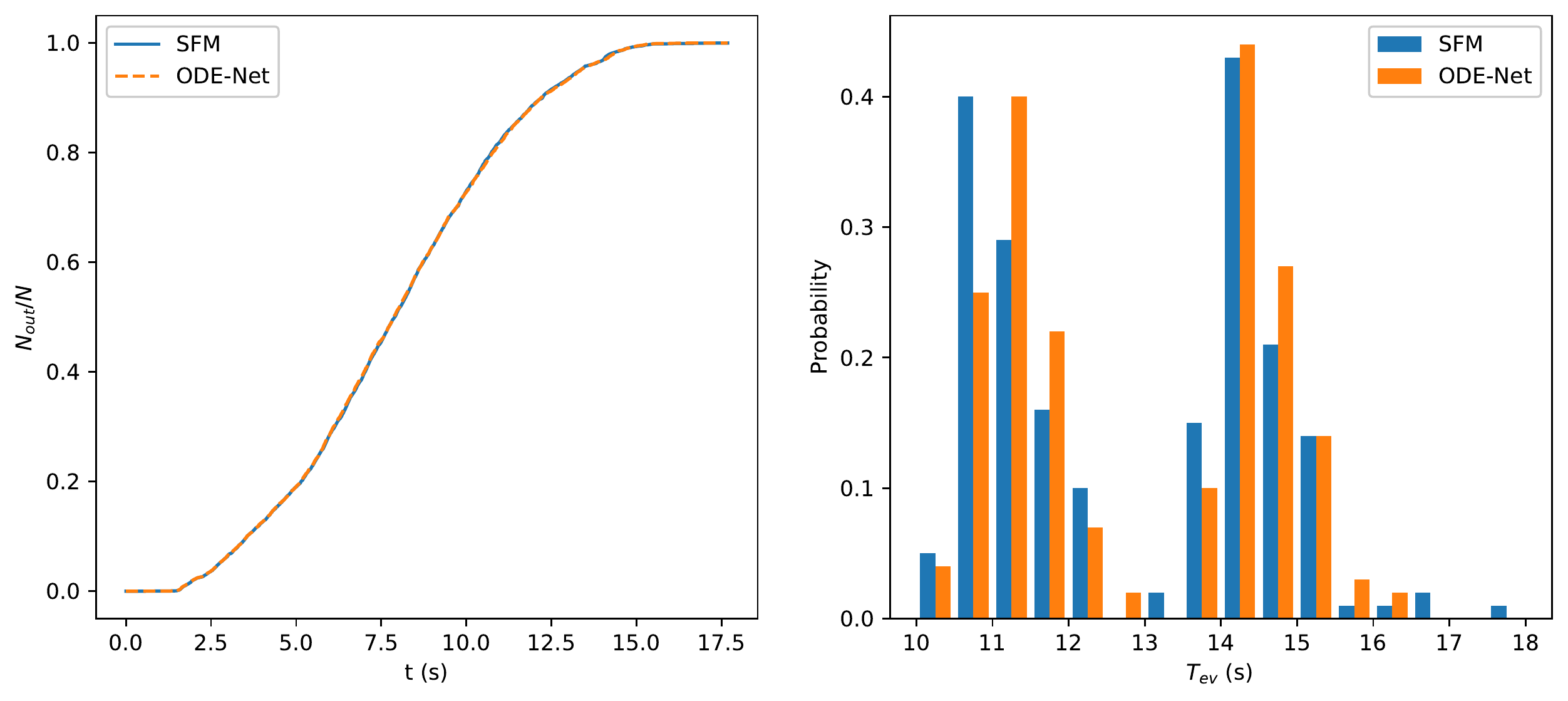}
\caption{Comparison of SFM and the learned ODE-Net model, for a crowd of 20 pedestrians. Left: ICE plotted against time, averaged over 200 simulations. Right: the histograms of the evacuation time $T_{ev}$ obtained from 200 simulations
of both models.}
\label{fig.hist_sfm20}
\end{figure*}

We demonstrate such a comparison in Figure~\ref{fig.sfm5}: the left figure shows the trajectories of the particles predicted by the actual SFM and the learned ODE-Net model, and the right one shows the particle's distance to the exit as a function of time, representing the velocity information of the particles. 
One can see from the figures that the results of the two models agree very well, indicating that ODE-Net can effectively learn the behaviors of the actual model in this case. We then test the learned ODE-Net model with a crowd of 20 particles, where the results are shown in Figure~\ref{fig.sfm20}. Once again we 
observe good agreement between the results of the two models,  suggesting that the ODE-Net trained 
with a crowd of 5 particles can be used to make predictions of a much larger crowd. 
That said, we do observe discrepancy in some trajectories near the exit,
which is likely due to the physical contacts between pedestrians when they are very close to each other, 
and such physical contacts are not taken into account in our present social force based model.  
Finally, as has been mentioned earlier, compared to the behaviors of each individual particles, it is more important to examine 
if the learned model can correctly predict important statistical or collective behaviors of the crowd, as
that is often what such models are used for. 
To this end, we consider the following two representative statistical quantities. 
First, we track the instantaneous collective escape (ICE) rate~\cite{D1SM00033K}, which is defined as the percentage of the pedestrians who have successfully exited the room
at a given time: $N_{out}/N$ where $N_{out}$ is the number of escaped pedestrians. 
We repeat the simulations of a crowd of 20 particles 200 times 
with random initial locations, and plot the average ICE rate 
as a function of time $t$ in Figure~\ref{fig.hist_sfm20} (left).
One can see that the result of SFM and that of the learned ODE-Net model look nearly identical. 
Another statistical quantity that we consider is the evacuation time $T_{ev}$, defined as the time for all the pedestrians to leave the room~\cite{D1SM00033K}.
We also perform 200 simulations with different initial locations, calculate the evacuation time for each simulation, and plot the histogram of it in
 Figure~\ref{fig.hist_sfm20}~(right). 
 Note that, the initial locations are chosen in a way
 that the resulting histogram is bi-modal,  to test if ODE-Net can capture this feature. 
 One can see from the figure that, the ODE-Net model does reproduce the bi-modal feature of 
 the histogram. 
 Finally it is important to note that, since the actual SFM is also based on an ODE system (therefore continues-time) and the social-force concept, 
 the ODE-Net performs very well in this experiment, thanks to the similarity between the actual model and ODE-Net. 
 To further test the ODE-Net method, next we apply it to a discrete-time model.

\subsection{Learning the ORCA model}
In this section, we apply ODE-Net to  the ORCA model, which is discrete in time and therefore 
conceptually different from  ODE-Net. 
As a discrete-time model, at each time step, ORCA allows each pedestrian to determine independently the optimal moving velocities and move accordingly~\cite{berg2011reciprocal}. Briefly speaking, the ORCA model assumes that each pedestrian can obtain the relative distance and velocity with respect to every neighboring pedestrian at a certain time step. Based on the information, the pedestrian computes a collision-free velocity for the next step of movement, by solving a constrained optimization problem. 
Specifically the velocity should be the one that is  closest to a prescribed target velocity,  
subject to the constraint that it will not cause collision with any other pedestrians or obstacles  in a finite time horizon. The details of the ORCA model can be found in~\cite{berg2011reciprocal} and the parameter values used in our simulation are presented in Table~\ref{tab.2}.

\begin{table}[!htb]
\small
\caption{List of parameter values in ORCA.}
\label{tab.2}
\begin{tabular}{lll}
 \hline
 Variable&Value&Description \\
 \hline
  $H_0$ & $10\,\rm{m}$ & side length of the room\\
  $N$ & $5(20)$ & number of pedestrians\\
  $r$ & $0.3\,\rm{m}$  & radius of pedestrians\\
  $v^p$ & $1.0\,\rm{m/s}$   & preferred velocity \\
  $\tau$ & $0.05\,\rm{s}$  & time horizon \\
  $\Delta t$ & $0.01\,\rm{s}$ & time step in simulation\\
  maxNeighbors & $10$ & max number of neighbors \\
  neighborDist & $2.5\,\rm{m}$  & max distance of neighbors \\
 \hline
\end{tabular}
\end{table}

\begin{figure*}[!htb]
\centering
\includegraphics[width=\textwidth]{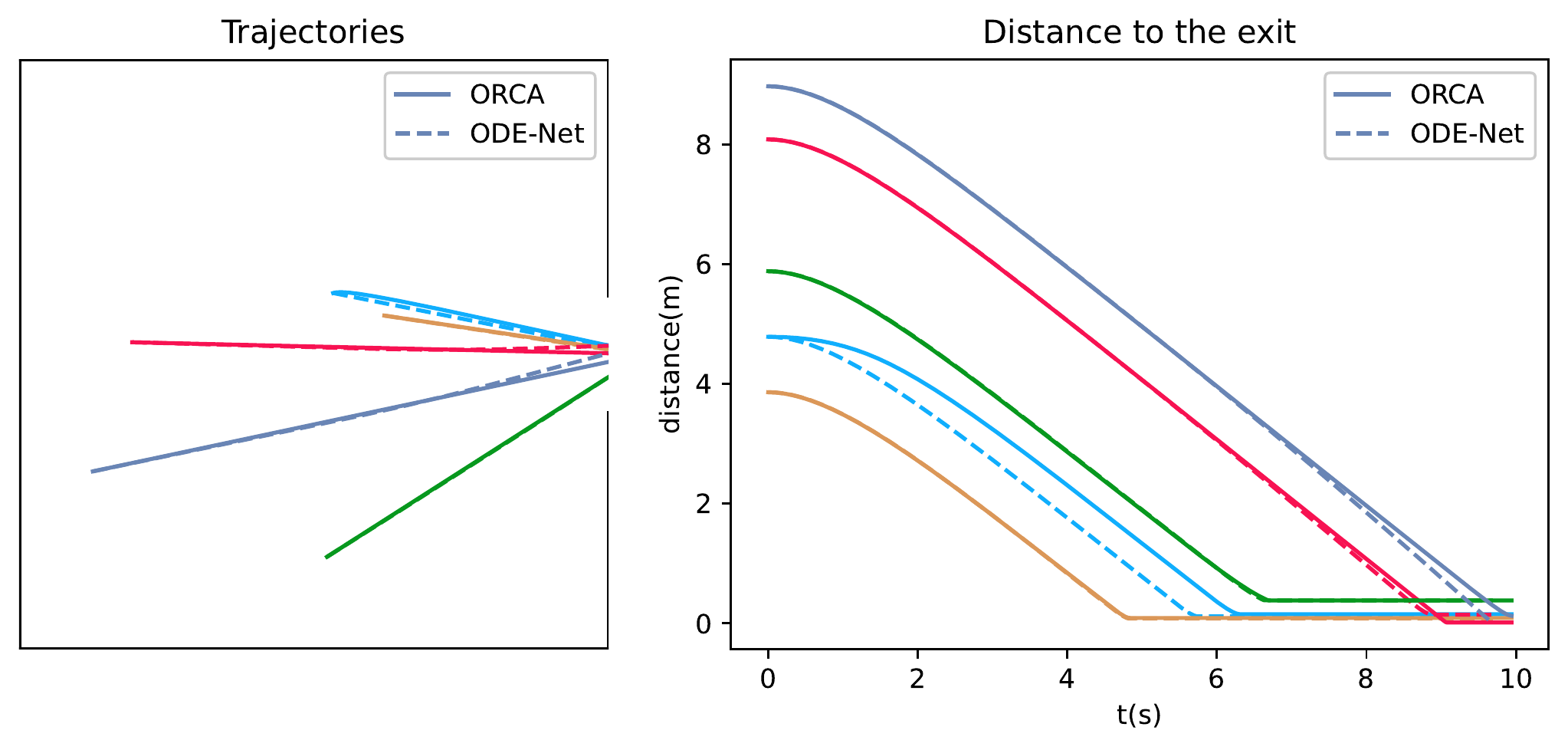}
\caption{Comparison of ORCA and the learned ODE-Net model, for a crowd of 5 pedestrians. Left: pedestrian trajectories generated with ORCA (solid) and ODE-Net (dotted). Right: each pedestrian's distance to the exit plotted against time,
where 
the results of ORCA and of ODE-Net are shown as solid and dotted lines respectively.}
\label{fig.orca5}
\end{figure*}

\begin{figure*}[!htb]
\centering
\includegraphics[width=\textwidth]{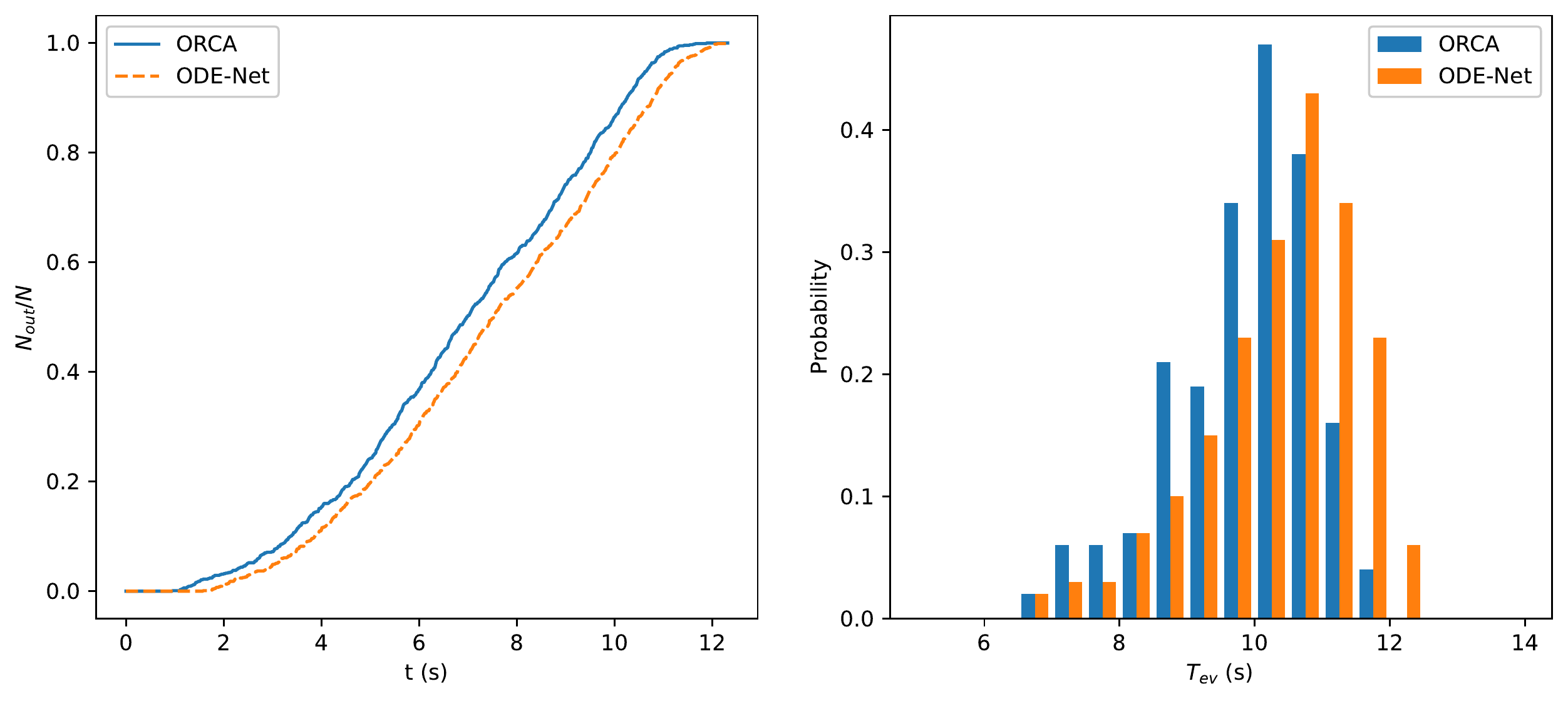}
\caption{Comparison of ORCA and the learned ODE-Net model, for a crowd of 5 pedestrians. Left: ICE plotted against time, averaged over 200 simulations. Right: the histograms of the evacuation time $T_{ev}$ obtained from 200 simulations
of both models.}
\label{fig.hist_orca5}
\end{figure*}

The training data are generated from the ORCA model for a crowd of 5 pedestrians,
and the training procedure is the same as that is used in the first example. 
As is in the first example, our first test is to apply the obtained ODE-Net model to a crowd of 5 pedestrians. 
Figure~\ref{fig.orca5} compares the trajectories and the distances to the exit predicted by both models,
in which one can see that the results agree quite well with each other. 
Next in Figure~\ref{fig.hist_orca5}, we show the average ICE rate and the histogram of the evacuation time, both obtained from 200 repeated trials. 
The plots illustrate that, although the discrepancies between the statistical results of the two models are larger than those in the first example,
the results largely agree with each other,  demonstrating the ability of the ODE-Net model to predict the 
crowd behaviors in this case. 

\begin{figure*}[!htb]
\centering
\includegraphics[width=\textwidth]{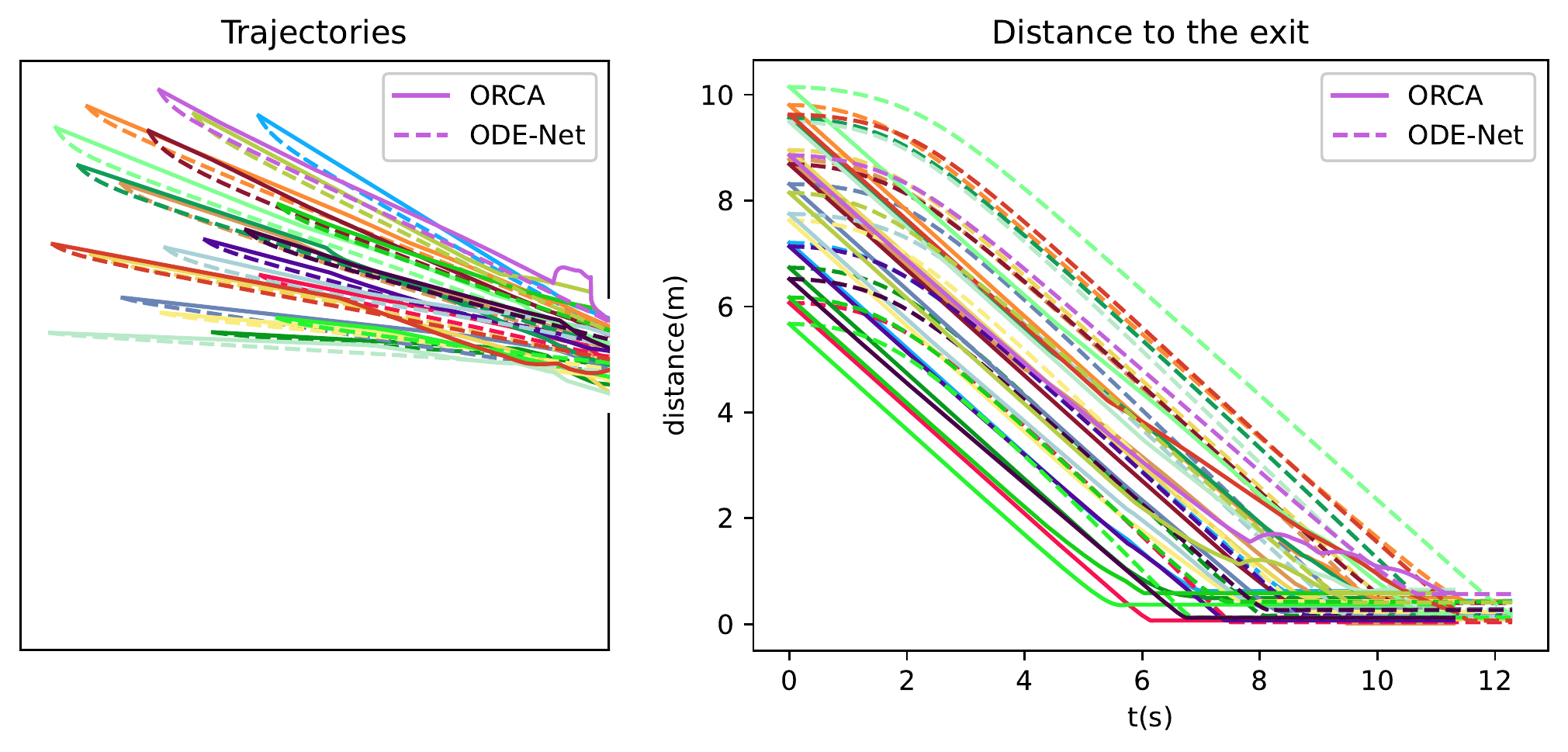}
\caption{Comparison of ORCA and the learned ODE-Net model, for a crowd of 20 pedestrians. Left: pedestrian trajectories generated with ORCA (solid) and ODE-Net (dotted). Right: each pedestrian's distance to the exit plotted against time,
where 
the results of ORCA and of ODE-Net are shown as solid and dotted lines respectively.}
\label{fig.orca20}
\end{figure*}

\begin{figure*}[!htb]
\centering
\includegraphics[width=\textwidth]{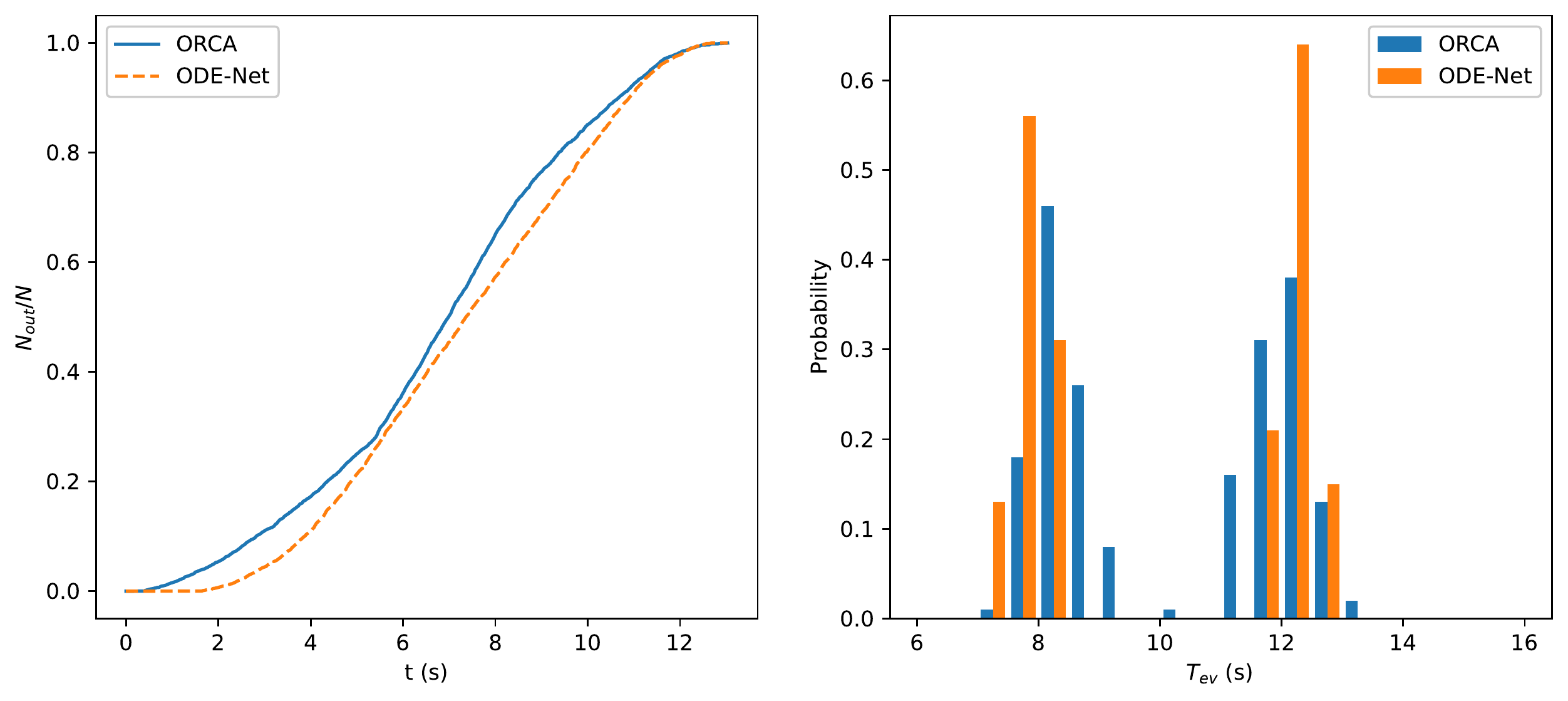}
\caption{Comparison of ORCA and the learned ODE-Net model, for a crowd of 20 pedestrians. Left: ICE plotted against time, averaged over 200 simulations. Right: the histograms of the evacuation time $T_{ev}$ obtained from 200 simulations
of both models.}
\label{fig.hist_orca20}
\end{figure*}

Next we consider a scenario that is more challenging for the obtained ODE-Net model, where we apply it to a crowd of 20 pedestrians that are closely spaced at the beginning. 
The particle trajectories and the distances to the exit are plotted  in Figure~\ref{fig.orca20},
where we see that the difference between the results of the two models become more substantial than that in the previous case, especially 
for the distance to the exit that represents the velocity information of the particles. 
Similar conclusions can be drawn from Figure~\ref{fig.hist_orca20}, which shows the ICE rate and the histogram of the evacuation time. 
In particular, we have found that while the ODE-Net captures the bimodal feature of the histogram, it seems to predict less variation in the results than the ORCA model. 
We believe that the larger discrepancy in this example is due to the fact that the ODE-Net and the ORCA models are different in nature:
first and foremost, one model is continuous in time and the other is discrete; moreover, how the interactions 
between particles take place in the two models is fundamentally different. 
We expect that the agreement can be improved by 
designing specific network structures according to the ORCA model, which is subject to further investigation. 

\section{Conclusion}
\label{sec:conclusion}
In this work we present an ODE-Net based method to learn continuous-time models
of crowd dynamics from data. 
In particular, we formulate the pedestrians as particles driven
by several socio-psychological forces, which are learned from the data.  
With the proposed method, we are able to learn models 
for large-scale and variable-size crowds, and 
use the learned model to make predictions 
for crowds that are larger or smaller than the training system. 
The performance of the proposed method is demonstrated by applying it to the synthetic data generated by two popular crowd dynamics models -- SFM and ORCA.
We believe that the proposed method can be useful in a range of applications, 
such as urban safety planning and human-robotic interaction.

Several potential improvements of the proposed method are possible. 
First, considering the limited view field of humans, interaction force between pedestrians might also depend 
on their facing directions. In particular, pedestrians would pay more attention to pedestrians in front of them than those behind them~\cite{johansson2007specification},
and such an effect should be taken into account when constructing the interaction force model. 
Second  we here assume that the personal motivation forces depend
only on the particle's velocity and position, which is certainly a simplification,
and as is discussed in Section~\ref{sec:sfm}, environmental factors should also be taken into account.
Finally, empirical studies show that a large fraction of people in a crowd move in small groups, such as friends walking together~\cite{moussaid2010walking},
the effect of which is not considered in the present work. To this end, it is desirable for the method to be able to  learn the group effects from data, 
and enhance the performance of the resulting model.
We plan to investigate these potential improvements in future studies.

\section*{Supplementary information} 
Animations of the crowd dynamics predicted by the learned ODE-Net and the true underlying models 
are provided in the supplementary material.
The code for the proposed method is available at \href{https://github.com/ChengChen0301/odenet\_crowd}{github.com/ChengChen0301/odenet\_crowd}.

% \bibliographystyle{mybib}
% \bibliography{refs}

\end{document}